\def\year{2020}\relax
\documentclass[letterpaper]{article} 
\usepackage{aaai20}  
\usepackage{times}  
\usepackage{helvet} 
\usepackage{courier}  
\usepackage[hyphens]{url}  
\usepackage{graphicx} 
\usepackage{graphicx}  
\usepackage{times}
\usepackage{soul}
\usepackage{url}
\usepackage{enumerate}
\usepackage{enumitem}
\usepackage[utf8]{inputenc}
\usepackage[small]{caption}
\usepackage{graphicx}
\usepackage{adjustbox}
\usepackage{amsfonts}
\usepackage{amssymb}
\usepackage{pifont}
\usepackage{amsmath}
\usepackage{booktabs}
\usepackage[caption = false]{subfig}
\usepackage{bibunits}
\usepackage[utf8]{inputenc}
\newcommand{\cmmnt}[1]{}

\usepackage{xcolor}
\usepackage{color, colortbl}
\usepackage{float}
\usepackage{xspace}
\usepackage{dblfloatfix} 
\usepackage{array}
\usepackage{arydshln}
\usepackage{algpseudocode}
\usepackage[ruled]{algorithm}
\usepackage[toc,page]{appendix}
\newcommand{\method}{\texttt{CUP}\xspace}
\newcommand{\methodss}{\texttt{CUP-SS}\xspace}

\definecolor{Gray}{gray}{0.9}

\newcommand{\M}[1]{\widetilde{\MakeUppercase{#1}}}
\newcommand{\V}[1]{\overline{\MakeUppercase{#1}}}
\newcommand{\SC}[1]{\MakeLowercase{#1}}

\newcommand{\cmark}{\ding{51}}%
\newcommand{\xmark}{\ding{55}}%

\title{\method: Cluster Pruning for Compressing Deep Neural Networks}

\author{Rahul Duggal,\textsuperscript{\rm 1}    Cao Xiao,\textsuperscript{\rm 2}    Richard Vuduc,\textsuperscript{\rm 1}    Jimeng Sun\textsuperscript{\rm 1}\\
\textsuperscript{\rm 1}Georgia Institute of Technology, Atlanta \\ \textsuperscript{\rm 2}Analytics Center of Excellence, IQVIA\\ 
rahulduggal@gatech.edu, cao.xiao@iqvia.com, richie@cc.gatech.edu, jsun@cc.gatech.edu,   
}

\begin{document}
 
\maketitle

\begin{abstract}
We propose Cluster Pruning (\method) for compressing and accelerating deep neural networks. Our approach prunes similar filters by clustering them based on features derived from both the incoming and outgoing weight connections.
With \method, we overcome two limitations of prior work---(1) \textbf{non-uniform pruning}: \method can efficiently determine the ideal number of filters to prune in each layer of a neural network. This is in contrast to prior methods that either prune all layers uniformly or otherwise use resource-intensive methods such as manual sensitivity analysis or reinforcement learning to determine the ideal number.
(2) \textbf{Single-shot operation}: We extend \method to \methodss (for \method single shot) whereby pruning is integrated into the initial training phase itself. This leads to large savings in training time compared to traditional pruning pipelines. 
Through extensive evaluation on multiple datasets (MNIST, CIFAR-10, and Imagenet) and models (VGG-16, Resnets-18/34/56) we show that \method outperforms recent state of the art. Specifically, \methodss achieves $2.2\times$ flops reduction for a Resnet-50 model trained on Imagenet while staying within $0.9\%$ top-5 accuracy. It saves over 14 hours in training time with respect to the original Resnet-50. Code to reproduce results is available here\footnote{\url{https://github.com/duggalrahul/CUP_Public}}.



\end{abstract}

\section{Introduction}

Deep neural networks (DNN) have achieved tremendous success in many application areas such as computer vision \cite{simonyan2014very},  language translation \cite{sutskever2014sequence} and disease diagnosis \cite{esteva2017dermatologist}. However, their performance often relies on billions of parameters and entails a large computational budget. For example, VGG-16 \cite{simonyan2014very} encumbers $16$ GFLOPS and $528$ MB of storage space for inference on a single image. This hinders their deployment in real-world applications that require low memory resources or entail strict latency requirements. Model compression aims at reducing the size, computation and inference time of a neural network while preserving its accuracy.

\begin{figure}[t]
\centering
\includegraphics[width = 0.45\textwidth]{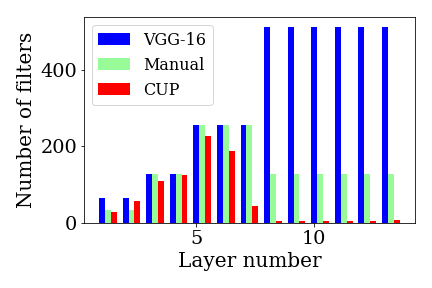}
\caption{Comparison on the number of filters in each layer of (1) the original VGG-16 model trained on CIFAR-10 (2) the compressed VGG-16 obtained through manual sensitivity analysis \protect\cite{li2016pruning} and (3) the compressed VGG-16 obtained through \method. Observe that the pruning pattern for \method correlates highly with the manual method. \method can automatically determine that layers 1 to 7 are more sensitive than layers 8 to 14. Consequently, the later layers are pruned more aggressively than the earlier ones.}
\label{Fig:crown_jewel}
\end{figure}

In this paper, we propose a new channel pruning based model compression method called {\it Cluster Pruning}  or \method, that clusters and prunes similar filters from each layer. One of the key challenges in channel pruning is to determine the optimal layerwise sparsity. For deep networks, introducing per layer sparsity as a hyperparameter leads to an intractable, exponential search space. 
Many prior works either completely avoid this problem by uniformly pruning all layers  \cite{he2017channel,he2018soft,he2019filter} or overcome it using costly heuristics such as manual sensitivity analysis \cite{li2016pruning}, reinforcement learning \cite{he2018amc}. Note that uniform pruning may lead to suboptimal compression since some layers are less sensitive to pruning than others \cite{liu2018rethinking}.

With \method, we enable layerwise \textbf{non-uniform pruning} whilst introducing only one hyperparameter $t$. In figure \ref{Fig:crown_jewel}, we compare the number of filters in the compressed VGG-16 model obtained via \method versus the model obtained through manual sensitivity analysis \cite{li2016pruning}. We observe that the pattern of pruning achieved by \method correlates highly with the manual method. Specifically, \method automatically determines that layers 1-7 are more sensitive to pruning than layers ($8-13$). Thus the later layers are pruned more aggressively than the earlier ones. Notice that the overall compression achieved by \method is significantly higher.

The second benefit of \method is that it can operate in the \textbf{single-shot setting} i.e. avoid any retraining whatsoever. This is achieved by extending \method to \methodss (for \method single shot) which iteratively prunes the original model in the initial training phase itself. This is done by calling \method before each training epoch. In contrast to recent methods \cite{he2018soft,he2019filter,zhao2019variational}, \methodss reduces the number of flops during training which results in large \textbf{savings in training time}. We compare the training times for different models on Imagenet in table \ref{tab:training_time_crown}. Notice that  \methodss saves over $14$ hours in training time while achieving $2.2\times$ flops reduction in comparison to training a full Resnet-50. The compressed model loses less than $1.5\%$ ($1\%$) in top-1 (top-5) accuracy compared to the original model. 
A substantial saving in training time can be huge for scenarios that involve frequent model re-training -- for example, healthcare wherein large models are retrained daily to account for data obtained from new patients.

\begin{table}[]
\begin{adjustbox}{width=\columnwidth,center}
\begin{tabular}{lccc} 
\toprule
\textbf{Model}   & \textbf{\begin{tabular}{c}Training \\ Time (hr) \end{tabular}} & \textbf{\begin{tabular}[c]{@{}c@{}}Flops\\ ($\times 10^9$)\end{tabular}} & \textbf{\begin{tabular}[c]{@{}c@{}}Accuracy (\%)\\ (Top-1 / Top-5)\end{tabular}} \\
\midrule
Resnet-50        & 66.00    & 8.21  & 75.87 \ / \ 92.87   \\
Resnet-50 \methodss & 51.58    & 3.72  & 74.39 \ / \ 91.94  \\                       
\bottomrule
\end{tabular}
\end{adjustbox}
\caption{Comparison of training time on Imagenet (ILSVRC 2012). Each model was trained on $3\times$ Titan Xp's with 12 GB RAM each. \methodss reduces the training time by over $14$ hours while reducing the flops by $2.2\times$. The final top-1/top-5 accuracy is within 1.5/1 percent of the original model.}
\label{tab:training_time_crown}
\end{table}

To summarise, in this paper we propose a new channel pruning method --- \method that has the following advantages: 
\begin{enumerate}[leftmargin=*]

    
    \item The pruning amount is controlled by a single hyperparameter. Further, for a fixed compression ratio, \method can automatically decide the appropriate number of filters to prune from each layer. 
    
    
    
    \item \method can operate in the single-shot setting (\methodss) thereby avoiding any retraining. This massively reduces the training time for obtaining the compressed model.
    
    
    \item Through extensive evaluations on multiple datasets (MNIST, CIFAR-10, and ImageNet) we show \method and \methodss provide state of the art compression performance. We open-source all code to enable reproduction and extension of the ideas proposed in this paper. 
    
    
\end{enumerate}
 
 The rest of the paper is structured as follows. In section \ref{sec:related_works}, we review related research. In section \ref{sec:method} we first describe the \method framework and then extend it to the single shot setting---\methodss. Then in section \ref{sec:experiments} we outline the evaluation metrics and evaluate our method against recent state of the art. In section \ref{sec:ablation_studies} we perform ablation studies and summarise our findings into actionable insights. Finally, in Section \ref{sec:conclusion} we conclude with a summary of our contributions.
 
\section{Related Work}
\label{sec:related_works}
Neural Network Compression is an active area of research wherein the goal is to reduce the memory, flops, or inference time of a neural network while retaining its performance. Prior work in this area can be broadly classified into the following four categories:

(1) \textit{Low-Rank Approximation}: where the idea is to replace weight matrices (for DNN) or tensors (for CNNs) with their low-rank approximations obtained via matrix or tensor factorization \cite{jaderberg2014speeding,wang2018wide}. (2) \textit{Quantization}  wherein compression is achieved by using lower precision (and fewer bits) to store weights and activations of a neural network \cite{courbariaux2015binaryconnect,jacob2018quantization}. (3) \textit{Knowledge Distillation} wherein compression is achieved by training a small neural network to mimic the output (and/or intermediate) activations of a large network \cite{hinton2015distilling,crowley2018moonshine}. (4) \textit{Pruning}  wherein compression is achieved by eliminating unimportant weights \cite{han2015learning,li2016pruning,zhuang2018discrimination}. Methods in these categories  are considered orthogonal and are often combined to achieve more compact models. 

Within pruning, methods can either be structured or unstructured. Although the latter usually achieves better pruning ratios, it seldom results in flops and inference time reduction. Within structured pruning, a promising research direction is channel pruning. Existing channel pruning algorithms primarily differ in the criterion used for identifying pruneable filters. Examples include pruning filters, with smallest L1 norm of incoming weights \cite{li2016pruning}, largest average percentage of zeros in their activation maps \cite{hu2016network}, using structured regularization \cite{wen2016learning,liu2017learning} or with least discriminative power \cite{zhuang2018discrimination}. Some of these methods \cite{li2016pruning,he2017channel} involve per layer hyperparameters to determine the number of filters to prune in each layer. However, this can be costly to estimate for deep networks. Other methods avoid this issue by doing uniform pruning \cite{he2018soft,he2019filter}. This, however, can be a restrictive setting since some layers are less sensitive to pruning than others \cite{li2016pruning}. 

Channel pruning based model compression has traditionally employed a three-step regime involving (1) training the full model (2) pruning the full model to desired sparsity and (3) retraining the pruned model. However, recently some works have tried to do away with the retraining phase altogether \cite{he2018soft,he2019filter,lin2019towards,zhao2019variational}. In this paper, we call this single-shot setting. Since it avoids any retraining, the single-shot setting can reduce the training time for obtaining a compressed model by atleast $2\times$. With \methodss, however, we are able to further reduce the training time by iteratively pruning the model during the training phase itself. The reduction in flops during training results manifests itself with a decrease in training time. In the next section, we describe the \method framework and discuss its extension to \methodss.

\section{\method Framework}
\label{sec:method}
\subsection{Notation}
\label{subsec:notation}
To maintain symbolic consistency, we use a capital symbol with a tilde, like $\M{W}$, to represent tensors of rank 2 or higher (i.e. matrices and beyond). A capital symbol with a bar, like $\V{F}$, represents a vector while a plain and small symbol like $b$ represents a scalar. A lower subscript indicates indexing, so, $\M{F}_{i,j}$ indicates the element at position $(i,j)$ while a superscript with parenthesis like $\M{W}^{(l)}$ denotes parameters for layer $l$ of the network.

\subsection{\method overview}
\label{subsec:method_overview}
\begin{figure}[h]
\centering
\includegraphics[width = 0.45\textwidth]{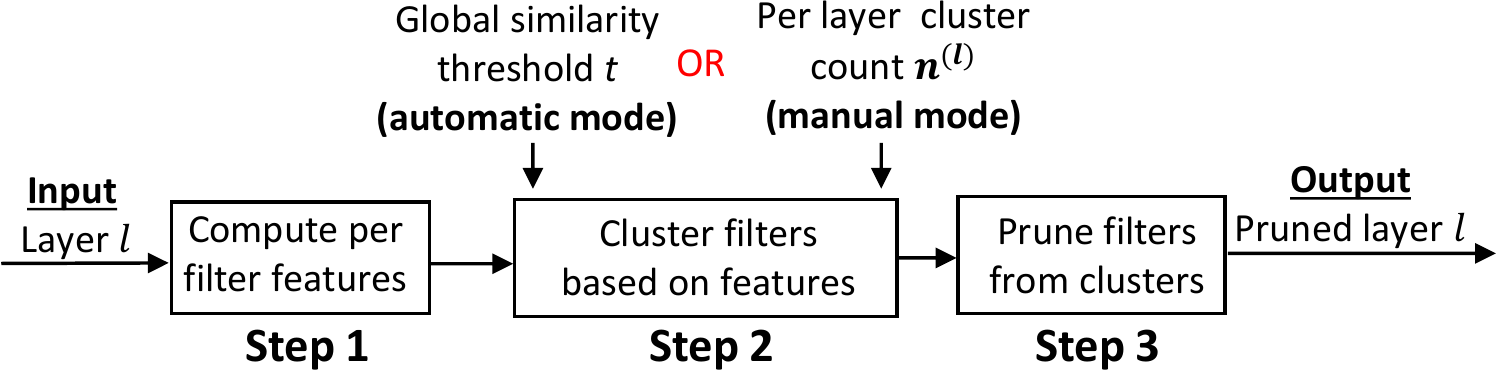}
\caption{Method overview for Cluster Pruning (\method).}
\label{Fig:method_overview}
\end{figure}
The \method pruning method is essentially a three-step process and is outlined in figure \ref{Fig:method_overview}. The first step computes features that characterize each filter. These features are specific to the layer type (fully connected vs convolutional) and are computed from the incoming and outgoing weight connections. The second step clusters similar filters based on the features computed previously. This step can operate in either manual or automatic mode. The third and last step chooses the representative filter from each cluster and prunes all others. In the following subsections, we go over the three steps in detail.

\subsection{Step 1 : Feature computation}
\label{subsec:step1_feature_computation}

The feature set for a filter depends on whether it is a part of a fully connected layer or a convolutional layer. In this section, we assume layers $l-1,l$ and $l+1$ of the neural network contain $n,m$ and $p$ filters respectively. 

\begin{figure}[h]
\centering
\subfloat[Fully connected layer features]{\includegraphics[width = 0.24\textwidth]{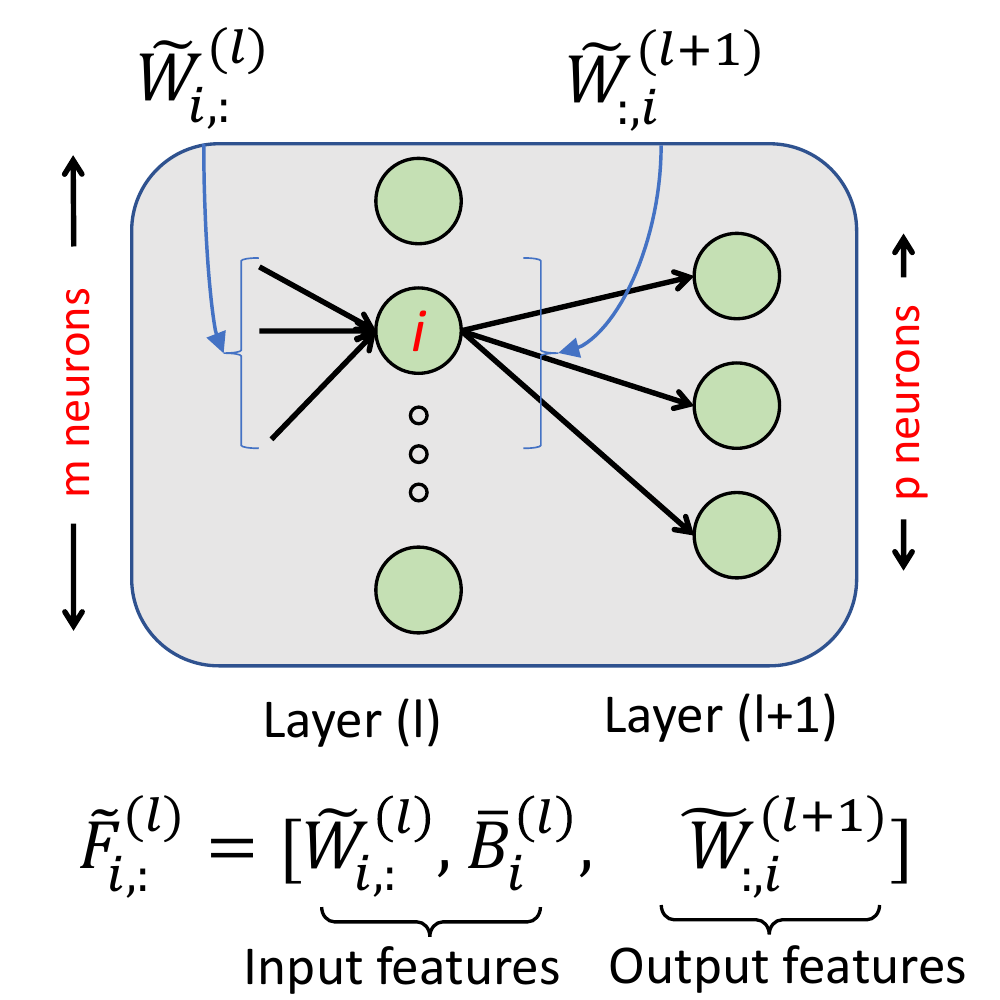} \label{fig:step_1_fullyconnected}} 
\subfloat[Convolutional layer features]{\includegraphics[width = 0.24\textwidth]{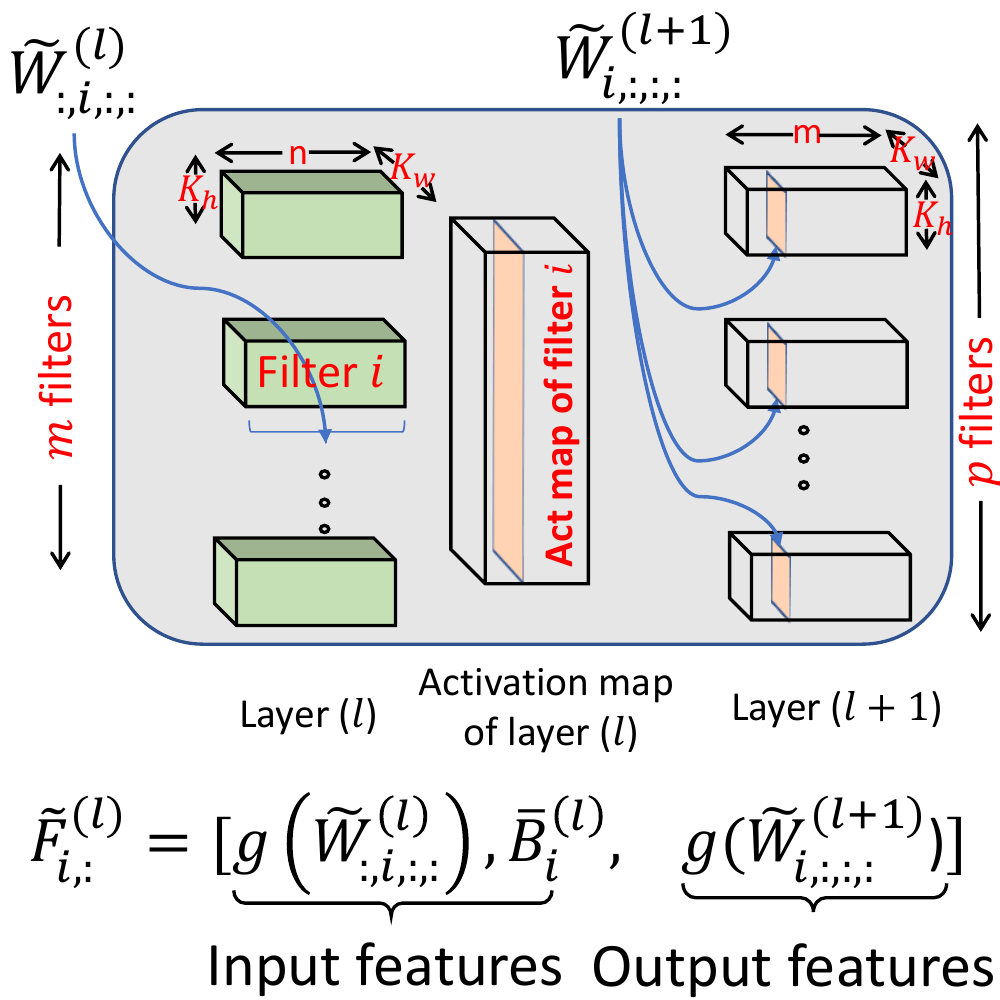} \label{fig:step_1_convolutional}} 
\caption{Feature computation for filters in (a) fully connected layers and (b) convolution layers of a neural network.}
\label{Fig:step_1_full}
\end{figure}

\subsubsection{\textbf{Fully Connected Layers}}:

The $l^{th}$ fully connected layer is parametrized by weights $\M{W}^{(l)}\in \mathbb{R}^{m\times n}$ and bias $\V{b}^{(l)} \in \mathbb{R}^m$. Within this layer, filter $i$ performs the following nonlinear transformation.
\begin{equation}
   \label{eq:fcn_op}
    \V{o}^{(l)}_i = \sigma(\M{W}^{(l)}_{i,:} \V{o}^{(l-1)} + \V{b}^{(l)}_i)
\end{equation} Where 
  $\V{o}^{(l-1)}$ is the output vector from layer $l-1$, $\V{W}^{(l)}_{i,:}, \V{b}^{(l)}_i$ are the $i^{th}$ row vector and element of the weight matrix and bias vector respectively. These also constitute the set of incoming weights to filter $i$. Finally, $\sigma$ is a non-linear function such as RELU. 
 Note that the weight vector $\V{W}^{(l+1)}_{:,i} \in \mathbb{R}^{p}$ corresponds to column $i$ of the weight matrix for layer $l+1$. Equivalently, these are also the weights on outgoing connections from filter $i$ in layer $l$.


We define $\M{F}^{(l)}_{i,:}$, the feature set of filter $i$ as

\begin{equation}
    \M{F}^{(l)}_{i,:} = \mathrm{concat}\underbrace{(\M{W}^{(l)}_{i,:},\V{b}^{(l)}_{i}}_\text{input features}, \underbrace{\M{W}^{(l+1)}_{:,i}}_\text{output features})
\end{equation}

where $\mathrm{concat}$ concatenates two vectors into one. We have $\M{F}^{(l)}_{i,:} \in R^{n+p+1}$ as shown in fig \ref{fig:step_1_fullyconnected}.




\subsubsection{\textbf{Convolutional Layers}}:
The $l^{th}$ convolutional layer is completely parameterized by the 4-D weight tensor $\M{W}^{(l)} \in R^{n \times m \times k_h \times k_w}$ and the bias vector $\V{b}^{(l)} \in \mathbb{R}^{m}$. The four dimensions of $\M{W}^{(l)}$ correspond to - number of input channels ($n$), number of filters in layer $l$ ($m$), height of filter ($k_h$) and width of filter ($k_w$).  The convolution operation performed by filter $i$ in layer $l$ is defined below

\begin{equation}
    \M{o}^{(l)}_{i,patch} = \sigma(\M{W}^{(l)}_{:,i,:,:} \otimes \M{o}^{(l)}_{patch} + \V{b}^{(l)}_i)
\end{equation}

Where $\M{o}^{(l)}_{i,patch} \in \mathbb{R}$ is the output of filter $i$ computed on a  $k_h \times k_w$ spatial crop $\M{O}^{(l)}_{patch} \in \mathbb{R}^{n \times k_h \times k_w}$ of the output map of layer $l-1$. Also $\M{W}^{(l)}_{:,i,:,:} \in \mathbb{R}^{n \times k_h \times k_w}$ and $\V{b}^{(l)}_i \in \mathbb{R}$ are the input weight tensor and bias for filter $i$ in layer $l$. $\otimes$ computes the elementwise dot product between 4-D tensors $A,B$ as $A \otimes B = \sum_a \sum_b \sum_c \sum_d A_{a,b,c,d}B_{a,b,c,d}$.




We define $\M{F}^{(l)}_{i,:}$, the feature set of filter $i$ as

 \begin{equation}
 \label{eq:conv_features}
     \M{F}^{(l)}_{i,:} = \mathrm{concat}(\underbrace{g(\M{W}^{(l)}_{:,i,:,:}),b^{(l)}_i}_\text{input features}, \underbrace{\M{W}^{(l+1)}_{i,:,:,:}}_\text{output features})
 \end{equation}
 
 
 \begin{equation}
     g(\M{X}_{:,:,:}) =[ \| \M{X}_{1,:,:}\|_F, \ldots, \| \M{X}_{C,:,:}\|_F]
 \end{equation}

Here $g: \mathbb{R}^{c} \times \mathbb{R}^{d} \times \mathbb{R}^{e} \rightarrow \mathbb{R}^c$ computes the channelwise frobenius norm of any arbitrary 3D tensor $\M{X}$. The input features in (\ref{eq:conv_features}) are computed from weights corresponding to filter $i$. The output features are computed from channels of filters in the succeeding layer that operate on the activation map produced by filter $i$. The input and output features are illustrated in figure \ref{fig:step_1_convolutional}. Like before, we have $\M{F}^{(l)}_{i,:} \in R^{n+p+1}$.


\subsection{Step 2: Clustering similar filters}
\label{subsec:step2_clustering}

Given feature vector $\M{F}^{(l)}_{i,:} \in \mathbb{R}^{n + p + 1}$ for each filter $i$ in layer $l$, we cluster similar filters in that layer using agglomerative hierarchical clustering \cite[Chapter~15]{wilks2011statistical}. The clustering algorithm permits two modes of operations - \textit{manual} and \textit{automatic}. These modes offer the tradeoff between 1) the user's control on the exact compressed model architecture and 2) The number of hyperparameters introduced in the compression scheme. 

\begin{itemize}
    \item \textbf{Manual mode}: This mode requires to specify the number of clusters $\SC{n}^{(l)}$ for each layer $l$. As we will see in step 3, $\SC{n}^{(l)}$ also equals the number of remaining filters in layer $l$ post pruning. This control over the exact architecture however, comes at the cost of expensive hyperparameter search to identify optimal $\SC{n}^{(l)}$. Prior work by \cite{he2018amc} demonstrates an automated strategy using reinforcement learning to successfully identify optimal $\SC{n}^{(l)}$. 
    
    \item \textbf{Automatic mode}: This mode uses a single hyperparameter - the global distance threshold $\SC{t}$ to automatically decide the number of filters to prune in each layer $l$. This mode trades the ability to specify exact models with introducing fewer hyperparameters - only 1 in this case. Automatic mode is advantageous in situations wherein hyperparameter search is costly. These might include pruning a deep model trained on large datasets where it may not be feasible to use an RL based search strategy to train 100s of candidate compressed models.
\end{itemize}
 
\begin{figure}[t]
\centering
\includegraphics[width = 0.5\textwidth]{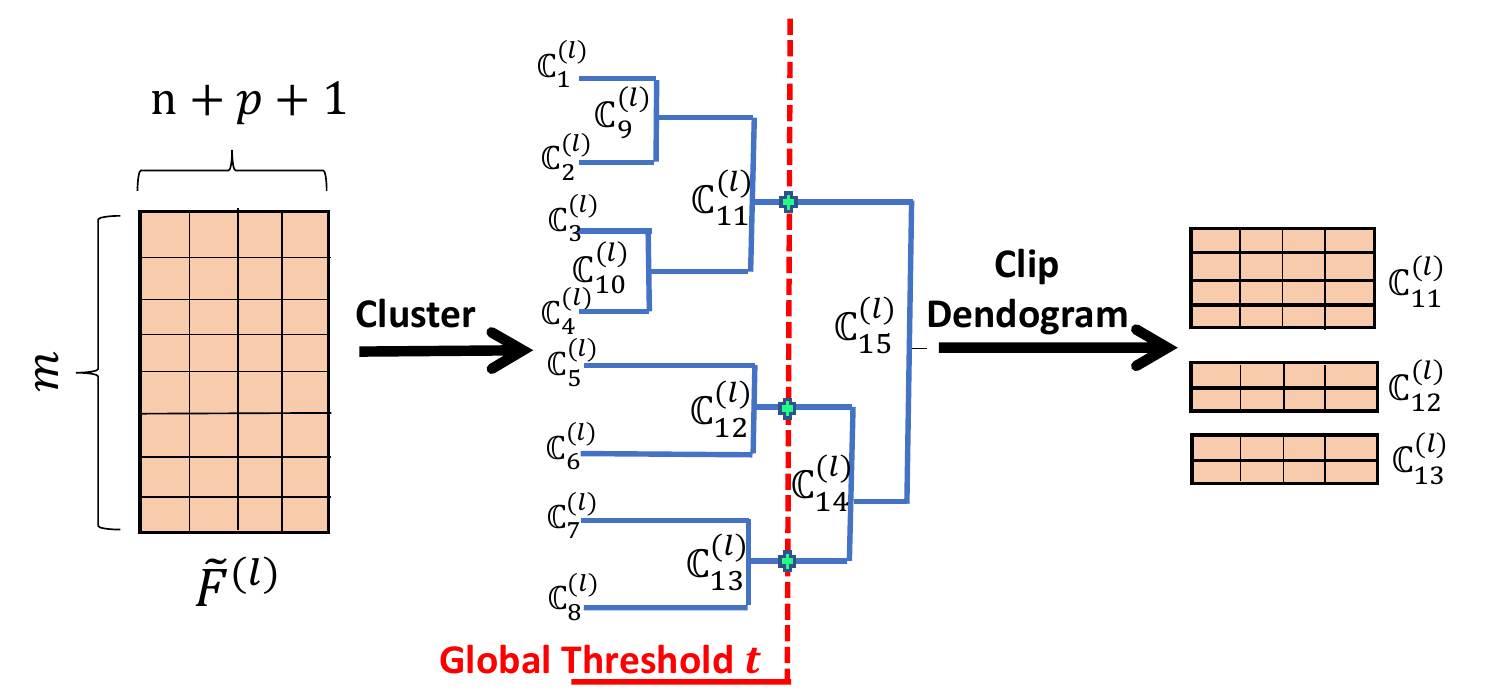}
\caption{Illustration for automatic mode clustering on layer $\SC{l}$ containing 8 filters. Clipping the dendrogram using threshold $\SC{t}$ leads to 3 clusters.}
\label{Fig:clustering_dendogram}
\end{figure}

 In our experiments, we use the standard python implementation of hierarchical clustering\footnote{\url{ https://docs.scipy.org/doc/scipy/reference/cluster.hierarchy.html}}. To cluster filters in layer $l$, the algorithm starts off with $m$ clusters $\mathbb{C}^{(l)}_i \forall i \in [1,m]$ each containing a single filter (Remember layer $\SC{l}$ contains $\SC{m}$ filters). In each iteration, the algorithm merges two clusters as per the \textit{Wards variance minimization} criterion. The criterion specifies to merge two clusters $\mathbb{C}^{(l)}_p,\mathbb{C}^{(l)}_q$ that lead to the least decrease in intra cluster variance over all possible pairings of clusters in $\mathbb{C}^{(l)}$. This clustering operation can be visualized as building a binary tree (called the dendrogram) with weighted edges as in figure \ref{Fig:clustering_dendogram}. Each non-leaf node of the dendrogram specifies a cluster with the height corresponding to the distance between its children (figure \ref{Fig:clustering_dendogram}). 
 
 The automatic mode corresponds to clipping the dendrogram at height $\SC{t}$, for all layers, thus leading to clusters corresponding to $\SC{t}$. Figure \ref{Fig:clustering_dendogram} illustrates automatic mode clustering for $8$ filters wherein the threshold $\SC{t}$ leads to $3$ clusters. In manual mode, the cluster merge operation is performed $m - \SC{n}^{(l)}$ times for layer $l$. Thus post clustering, we have exactly $\SC{n}^{(l)}$ clusters or $|\mathbb{C}^{(l)}| = \SC{n}^{(l)}$.  Regardless of operating mode, the output of step 2 is the set of clusters $\mathbb{C}^{(l)}$ corresponding to filters in layer $l$ of the neural network.

\subsection{Step 3 : Dropping filters from clusters}
\label{subsec:step3_pruning}
Given the set of filter clusters $\mathbb{C}^{(l)}$ for layer $l$, we formulate pruning as a subset selection problem. The idea is to determine the most representative subset of filters $\mathbb{S}^{(l)}_r$ for all filter clusters $\mathbb{C}^{(l)}_r \in \mathbb{C}^{(l)}$. Pruning then corresponds to replacing all filters in $\mathbb{C}^{(l)}_r$ by $\mathbb{S}^{(l)}_r$. Note that $\mathbb{S}^{(l)}_r$ also denotes the set of filters remaining after pruning.

Motivated by prior work, several subset selection criterion can be formulated as below
\begin{itemize}
    \item \textbf{Norm based} :  \cite{li2016pruning} prune filters based on the $l1$ norm of incoming weights. So norm based subset selection may amount to choosing the top $k\%$ filters having the highest feature norm as the cluster representative.
    \item \textbf{Zero activation based} : \cite{hu2016network} prune filters based on the average percentage of zeros (ApoZ) in their activation map when evaluate over a held-out set. So this method may amount to choosing the top $k\%$ filters having least ApoZ as the cluster representative.
    \item \textbf{Activation reconstruction based} : \cite{he2017channel} prunes filters based on its contribution towards the next layer's activation. So this method may amount to choosing top $k\%$ filters having maximum contribution.
\end{itemize}


In this paper, we chose the norm based subset selection criterion while constraining the cluster representative to contain a single filter. Thus

\begin{equation}
    \mathbb{S}^{(l)}_r = \underset{i \in \mathbb{C}^{(l)}_r}{\text{argmax}}\|\M{F}^{(l)}_{i,:}\|_2
\end{equation}

Here $\M{F}^{(l)}_{i,:} \in \mathbb{R}^{n+p+1}$ is the feature set for f $i$ containing both the input and output features. Note that since $|\mathbb{S}^{(l)}_r|=1$, number of remaining filters in layer $\SC{l}$ post pruning equals the number of clusters.

\subsection{Single-shot pruning}
\label{subsec:single_shot}
The traditional pruning pipeline involves a three-step process - 1) train a base model, 2) prune to desired sparsity and 3) retrain. In the single shot pruning, steps 2,3 are discarded and pruning is performed during the initial training phase. We enable single-shot pruning through \methodss whereby a small portion of the original model is pruned after every epoch of the initial training phase. This is done by calling \method with a linearly increasing value for $\SC{t}$ in each epoch of the training phase. We find that the following linear schedule for $t(e)$ (value of $t$ at epoch $e$) suffices for good performance. 

\begin{equation}
    t(e) = k . e + b
    \label{eq:single_shot_rate}
\end{equation}

Here $\SC{k},\SC{b}$ are hyperparameters controlling the slope and offset of the linear pruning schedule. A direct consequence of single-shot pruning is a large reduction in training time. The \methodss algorithm is outlined in algorithm \ref{algo1:cupss}.

\begin{algorithm}[h!]
\caption{\methodss : Single Shot Cluster Pruning}
\label{algo1:cupss}
\begin{algorithmic}[1]
\Procedure{CUPSS}{$model,tgt\_flops,k,b$}   
    \State $e \leftarrow 0$
    \State $t \leftarrow 0$
    \While{$e \not= n\_epochs$}  
        \State $e \leftarrow e+1$  
        \If{$get\_flops(model) \geq tgt\_flops$}
            \State $t \leftarrow k.e + b$          \Comment{e.q. (\ref{eq:single_shot_rate})}
            \State $model \leftarrow CUP(model,t)$ 
        \EndIf
        \State $model \leftarrow TRAIN(model,n\_epochs=1)$
    \EndWhile
\EndProcedure
\end{algorithmic}
\end{algorithm}

\section{Experiments}
\label{sec:experiments}
\subsection{Datasets}
\label{subsec:datasets}
We evaluate our methods against prior art on three datasets.
\begin{itemize}
    \item \textbf{MNIST} \cite{lecun1998gradient} : This dataset consists of 60,000 training and 10,000 validation images of handwritten digits between 0-9. The images are greyscale and of spatial size $28 \times 28$. 
    \item \textbf{CIFAR-10} \cite{krizhevsky2009learning} : These datasets consist of 50,000 training and 10,000 validation images from 10 classes. The images are 3 channel RGB, of spatial size $32 \times 32$.
    \item \textbf{Imagenet 2012} \cite{russakovsky2015imagenet} : This dataset consists of 1.28 million images from 1000 classes in the training set and 50,000 validation images. The images are 3 channel RGB and are rescaled to spatial size $224 \times 224$.
\end{itemize}

\subsection{\textbf{Training Procedure \& Baseline Models}}
\label{subsec:training_hyperparam_and_baselines}
For MNIST and CIFAR datasets, we reuse the training hyperparameter settings from \cite{liu2017learning}. All networks are trained using SGD with batch size - $64$, weight decay - $10^{-4}$, initial learning rate - $0.1$ which is decreased by a tenth at $1/2$ and $3/4$ the number of total epochs. On MNIST, we train for 30 epochs while on CIFAR 10 the networks are trained for 160 epochs. On Imagenet, we train with a batch size of $256$ for $90$ epochs. The initial learning is set to $0.1$. This is reduced by a tenth at epochs $30$ and $60$. We use a weight decay of $10^{-4}$. After compression, the model is retrained with the same strategy. The baseline models are described next.

\begin{itemize}
    \item \textbf{ANN}: We use the multiple layer perceptron with 784-500-300-10 filter architecture  in \cite{liu2017learning,wen2016learning}. This is trained upto $98.63\%$ accuracy.
    \item \textbf{VGG-16} \cite{simonyan2014very}: This the standard VGG architecture with batchnorm layers after each convolution layer. It is trained up to $93.64\%$ on CIFAR-10.
    \item \textbf{Resnet-56} \cite{he2016deep}: We use the standard Resnet model which is trained up to $93.67\%$ on CIFAR-10.
    \item \textbf{Resnet-\{18,34,50\}} \cite{he2016deep}: We use the official pytorch implementations for the three models. The baselines models have $69.87\%$, $73.59\%$ and $75.86\%$ Top-1 accuracies on Imagenet respectively. For Resnets, we only prune the first layer within each bottleneck.
\end{itemize}

\subsection{Comparison metrics}
\label{subsec:comparison_metrics}
We benchmark \method against several recent state of the art compression methods. Whenever possible, relevant results are quoted directly from the referenced paper. For \cite{wen2016learning} we use our implementation. The compression metrics are reported as $M = \frac{P_{base}}{P_{compressed}}$ where $P$ can be one of the following measured attributes (MA). 

\begin{itemize}
    \item Parameter reduction (PR) : The MA is number of weights.
    \item Flops reduction (FR) : The MA is number of multiply and add operations to score one input.
    \item Speedup: The MA is time to score one input on the CPU.
\end{itemize}

\begin{table}[b!]
\begin{adjustbox}{width=\columnwidth,center}
\begin{tabular}{@{}lccc@{}}
\toprule
\textbf{Method}             & \textbf{Model}   & \multicolumn{2}{c}{\textbf{Accuracy change} (\%)}  \\ \midrule
                   &                & \begin{tabular}[c]{@{}c@{}}Without\\  retraining\end{tabular} & \begin{tabular}[c]{@{}c@{}}With \\ retraining\end{tabular} \\
Base               & B & 0             & 0         \\
random             & M2  & -57.04\%    & -0.18\%   \\
L2                 & M2  & -19.04\%    & -0.16\%    \\
L1 \cite{li2016pruning}& M2  & -18.48\%  & -0.25\%   \\
SSL \cite{wen2016learning}   & M1  & -   & -0.10\%     \\
Slimming \cite{liu2017learning} & M2  & -    & -0.06\%     \\
\rowcolor{Gray}
\method    & M2  & \textbf{-13.26\%}                                            & \textbf{0.00\%}                               \\ \bottomrule
\end{tabular}
\end{adjustbox}
\caption{Accuracy change while compressing base model B ($784-500-300-10$) to the compressed model M1 ($434-174-78-10$) and M2 ($784-100-60-10$). \method achieves the best accuracy with and without retraining. }
\label{tab:MNIST}
\end{table}

\subsection{Comparison on accuracy change}
\label{subsec:comparison_on_accuracy_change}
\noindent \textbf{Goal}: To measure the change in accuracy while compressing model B to M2 on the MNIST dataset. 

\noindent Here, model B is a four layer fully connected neural network and consists of 784-500-300-10 filters \cite{wen2016learning,liu2017learning}. In \cite{wen2016learning} model B is compressed to model M1: 434-174-78-10 which corresponds to $83.5\%$ parameter reduction. In \cite{liu2017learning} it is compressed to model M2: $784-100-60-10$ which corresponds to $84.4\%$ sparsity. 

\noindent \textbf{Result}: We present results in table \ref{tab:MNIST} under two settings---(1) ``Without retraining'' wherein the trained model is pruned and no retraining is done thereafter. A lesser drop in accuracy signifies the robustness of our \textit{filter saliency} criterion. For further intuition on our filter saliency criterion and its connection to magnitude pruning, we refer the reader to Appendix \ref{appendix:saliency_and_connection}. Under setting (2): ``with retraining'', we retrain the model post pruning. Observe that \method fully recovers model B's performance and there is no loss in the compressed model's accuracy.

\noindent \textbf{Key Takeaway}: \method leads to the smallest drop in accuracy.

\begin{table}[t!]
\centering
\begin{adjustbox}{width=\columnwidth,center}
\begin{tabular} 
 {>{\raggedright\arraybackslash}p{4.6cm}%
  >{\raggedright\arraybackslash}p{0.2cm}%
  >{\centering\arraybackslash}p{2.9cm}%
  >{\raggedright\arraybackslash}p{1.1cm}%
  }


\toprule
\textbf{Method}        & \begin{tabular}{@{}l@{}}\textbf{Single} \\ \textbf{shot} \end{tabular} & \textbf{FR} ($\times$) & \begin{tabular}[l]{@{}l@{}}\textbf{Accuracy} \\ \textbf{Change} \end{tabular} \\ \midrule 
\multicolumn{4}{c}{\textbf{Resnet-56 on CIFAR 10}} \\ \hline
L1 \cite{li2016pruning}      & \xmark     & $1.37\times$   & -0.02\%        \\
ThiNet \cite{luo2017thinet}   & \xmark     & $1.99\times$   & -0.82\%       \\
CP \cite{he2017channel}     & \xmark      & $2.00\times$      & -1.00\%           \\
GM \cite{he2019filter}   & \xmark   & $2.10\times$    & -0.33\% \\
GAL-0.8 \cite{lin2019towards}    & \xmark   & $2.45\times$    & -1.68\% \\
\rowcolor{Gray}
\method \cmmnt{(t=0.8)}          & \xmark  &    $\textbf{2.77}\times$   & -0.40\%    \\ \hdashline[3pt/3pt]
SFP \cite{he2018soft}   & \cmark   & $2.10\times$    & -1.33\% \\
VCNP \cite{zhao2019variational}  &  \cmark   & $1.25\times$    & -0.78\% \\
GAL-0.6 \cite{lin2019towards} & \cmark & $1.59\times$  & -$0.28\%$                      \\
GM \cite{he2019filter}   & \cmark   & $2.10\times$    & -0.70\% \\ 
\rowcolor{Gray}
\methodss \cmmnt{(k=0.02, b=0)}       & \cmark  &  $\textbf{2.12}\times$  & -\textbf{0.31}\% \\ \midrule

\multicolumn{4}{c}{\textbf{VGG-16 on CIFAR 10}} \\ \hline
&  & \textbf{PR} ($\times$) / \textbf{FR} ($\times$) &  \\ 
L1 \cite{li2016pruning}      & \xmark  & $2.77\times$   / $1.51\times$   & -0.15\%       \\
ThiNet \cite{luo2017thinet}  &  \xmark  & $1.92\times$   / $2.00\times$      & -0.14\%   \\
CP \cite{he2017channel}    &  \xmark  & $1.92\times$   / $2.00\times$      & -0.32\%     \\
Slimming \cite{liu2017learning} & \xmark & $8.71\times$   / $2.04\times$   & -0.19\%     \\
VIB \cite{dai2018compressing}& \xmark & $17.27\times$  / $3.43\times$   & -0.20\%                                                     \\
GAL-0.1 \cite{lin2019towards}& \xmark & $5.61\times$  / $1.82\times$   & -$0.54\%$                      \\
\rowcolor{Gray}
\method \cmmnt{(t=0.95)}     &  \xmark   & $\textbf{23.38}\times$  / $\textbf{3.70}\times$     & -0.70\% \\ \hdashline[3pt/3pt]
VCNP  \cite{zhao2019variational} & \cmark   & $3.75\times$  / $1.64 \times$                & -0.07\% \\
GAL-0.1 \cite{lin2019towards}& \cmark & $5.61\times$  / $1.82\times$   & -$3.18\%$                      \\
\rowcolor{Gray}
\methodss \cmmnt{(k=0.035, b=0)}   &  \cmark     & $\textbf{17.12}\times$  / $\textbf{3.15}\times$ & -0.40\% \\


\bottomrule
\end{tabular}
\end{adjustbox}
\caption{Comparison with prior art on CIFAR-10. The dotted line separates methods with and without re-training. FR and PR indicate the flops and parameter reduction (higher is better). Our goal here is to maximize FR and PR for an acceptable change in accuracy. }
\label{Tab:cifar10_results}
\end{table}

\subsection{Comparison on parameter and flop reduction}
\label{subsec:comparison_on_param_and_flop_reduction}
\noindent\textbf{Goal}: To maximize the flops (or parameter) reduction while maintaining a reasonable drop in accuracy.

\noindent \textbf{Result}: In tables \ref{Tab:cifar10_results}, \ref{Tab:imagenet_results} we present the flops and parameter reduction along with the accuracy change from the baseline model. For a fair comparison, we consider two settings - ``With retraining'' which is indicated by a \xmark, and ``without retraining'' or single shot which is indicated by a \cmark. 

\textbf{With retraining}: In the first half of tables \ref{Tab:cifar10_results}, \ref{Tab:imagenet_results}, we consider the traditional 3 stage pruning setting where retraining is allowed. As a model designer using \method, our aim is to set as high a $t$ that leads to some acceptable loss in accuracy. Consequently, we identify the optimal $t$ using a simple linesearch as detailed in the following section on ablation study and Appendix \ref{appendix:linesearch}. From the tables we see that \method always leads to a higher flops reduction (FR) for a lower or comparable drop in accuracy.

\textbf{Without retraining}: In the lower half of tables \ref{Tab:cifar10_results}, \ref{Tab:imagenet_results}, we don't allow any retraining post pruning. As a model designer using \methodss, our aim is to identify the optimal $k,b$ that lead to some acceptable loss in accuracy. We determine the best $k,b$ through a line search on $k$ while constraining $b=0$ (for CIFAR-10) and $b=0.3$ (for Imagenet). Detailed analysis is presented in the section on ablation study and Appendix \ref{appendix:linesearch}.

\noindent \textbf{Key Takeaway}: We notice that when retraining is allowed, we always obtain higher compression regardless of the method used. This is to be expected since multi-shot methods expend more resources while training. The real benefit for single-shot methods comes with the drastic reduction in training time as we will see in the next subsection.

\begin{table}[t!]
\centering
\begin{adjustbox}{width=\columnwidth,center}
\begin{tabular}{p{4.5cm}lrr}
\toprule
\textbf{method} & \begin{tabular}{@{}l@{}}\textbf{Single} \\ \textbf{shot} \end{tabular}& \textbf{FR} ($\times$) & \textbf{Accuracy change} \\ \midrule

\multicolumn{4}{c}{\textbf{Resnet-18 on ImageNet}} \\ 
\midrule
                  &        &          & \textbf{Top1} \  \ / \ \ \textbf{Top 5}             \\
Slimming \cite{liu2017learning}       &   \xmark    & $1.39\times$             & -1.77\% \ /  \  -1.29\%       \\
GM \cite{he2019filter}  &  \xmark    & $1.71\times$             & -1.87\% \ / \ -1.15\%               \\
COP \cite{wang2019cop}   &   \xmark    & $1.75\times$             & -2.48\% \ \ / \ \ \ \ \ NA \ \              \\ 
\rowcolor{Gray}
\method \cmmnt{(t=0.80)}   &  \xmark  & $\textbf{1.75}\times$             & \textbf{-1.00}\% \ / \ \textbf{-0.79}\%               \\  \hdashline[3pt/3pt]
SFP \cite{he2018soft}   &   \cmark    & $1.71\times$             & -3.18\% \ / \ -1.85\%               \\
GM \cite{he2019filter}  &  \cmark    & $1.71\times$             & -\textbf{2.47}\% \ / \ -1.52\%               \\  
\rowcolor{Gray}
\methodss \cmmnt{(k=0.03, b=0.3)}   & \cmark  &  $\textbf{1.75}\times$  & -2.50\% \ / \ -\textbf{1.40}\% \\ \midrule

\multicolumn{4}{c}{\textbf{Resnet-34 on ImageNet}} \\ 
\midrule
                  &     &     \textbf{FR} ($\times$)        & \textbf{Top1} \  \ / \ \ \textbf{Top 5}             \\

L1 \cite{li2016pruning}   &    \xmark   &  $1.31 \times$            &     -1.06\% \ \ / \ \ \ \ NA \ \                        \\
GM \cite{he2019filter}    &   \xmark     & $1.69\times$             & -1.29\% \ / \ -0.54\%               \\
\rowcolor{Gray}
\method \cmmnt{(t=0.60)}  & \xmark  & $\textbf{1.78}\times$             & -\textbf{0.86}\% \ / \ -\textbf{0.53}\%              \\ \hdashline[3pt/3pt]
SFP \cite{he2018soft}    &   \cmark       & $1.69\times$             & -2.09\% \ / \ -1.29\%               \\
GM \cite{he2019filter}    &   \cmark     & $1.69\times$             & -2.13\% \ / \ -\textbf{0.92}\%               \\
\rowcolor{Gray}
\methodss \cmmnt{(k=0.03, b=0.3)}   & \cmark  &  $\textbf{1.71}\times$              & -\textbf{1.61}\% \ / \ -1.02\% \\ \midrule

\multicolumn{4}{c}{\textbf{Resnet-50 on ImageNet}} \\ 
\midrule
        &          &        \textbf{FR} ($\times$)          & \textbf{Top1} \  \ / \ \ \textbf{Top 5}             \\

Thinet \cite{luo2017thinet}    &   \xmark    & $2.25\times$             & -1.87\% \ / \ -1.12\%              \\
CP  \cite{he2017channel}      &   \xmark     & $2.00\times$                & NA \ / \ -1.40\%                   \\
NISP \cite{yu2018nisp}    &  \xmark   & $1.78\times$             & -0.89\% \ \ / \ \ \ \ \ NA \ \                 \\
SFP \cite{he2018soft}      &   \xmark     & $2.15\times$             & -14.0\% \ / \ -8.20\%       \\ 
FCF \cite{li2019compressing}      &    \xmark    & $2.33\times$             & -1.60\% \ / \ -0.69\%               \\
GAL-0.5-joint \cite{lin2019towards}      &    \xmark    & $2.22\times$             & -4.35\% \ / \ -2.05\%               \\
GM \cite{he2019filter}      &    \xmark    & $2.15\times$             & -1.32\% \ / \ -\textbf{0.55}\%               \\
\rowcolor{Gray}
\method \cmmnt{(t=0.7)}   &  \xmark  & $\textbf{2.47}\times$             & -1.27\% \ / \ -0.81\%               \\ \hdashline[3pt/3pt]
SFP \cite{he2018soft}     &  \cmark      & $1.71\times$             & -1.54\% \ / \ -\textbf{0.81}\%       \\
GM \cite{he2019filter}      &    \cmark    & $2.15\times$             & -2.02\% \ / \ -0.93\%               \\ 
\rowcolor{Gray}
\methodss \cmmnt{(k=0.03, b=0.3)}   & \cmark  &  \textbf{2.20}$\times$  & -\textbf{1.47}\% \ /  \  -0.88\% \\ \bottomrule
\end{tabular}
\end{adjustbox}
\caption{Comparison with prior art using Resnets trained and evaluated on Imagenet. FR indicates flop reduction (higher is better). NA denotes non-availability in the original paper.  Notice that some methods achieve lesser accuracy change. However it must be noted that these models are less compact as evidenced by the lower FR. }
\label{Tab:imagenet_results}
\end{table}

\begin{figure*}[t!]
\centering
\subfloat[]{\includegraphics[width = 0.25\textwidth]{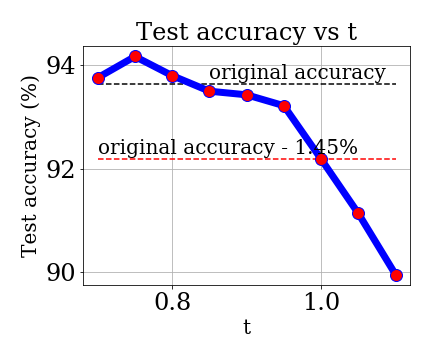} \label{Fig:acccuracy_vs_T}} 
\subfloat[]{\includegraphics[width = 0.25\textwidth]{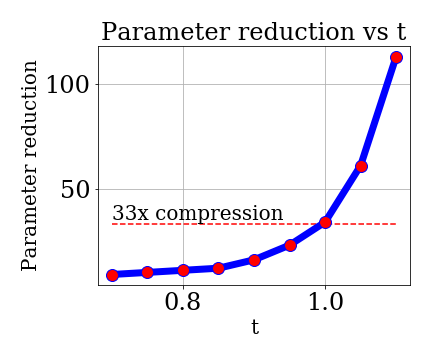} \label{Fig:parameter_reduction_vs_T}} 
\subfloat[]{\includegraphics[width = 0.25\textwidth]{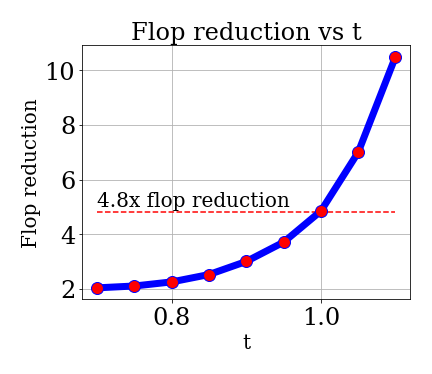}  \label{Fig:flop_reduction_vs_T}} 
\subfloat[]{\includegraphics[width = 0.25\textwidth]{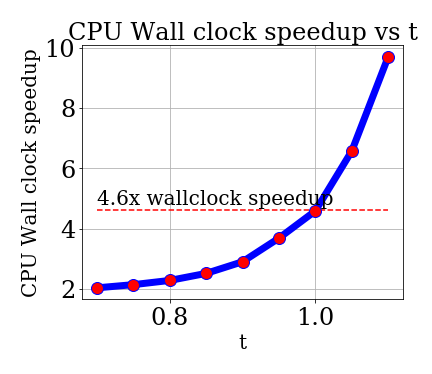} \label{Fig:wallclock_reduction_vs_T}} 
\caption{Results for compressing VGG-16 (on CIFAR-10) versus global threshold $t$. (a) accuracy of the compressed model (in \%) versus  $t$ (b) parameter reduction versus $t$, (c) flop reduction versus $t$, (d) shows wall clock speedup versus $t$.}
\label{Fig:T_analysis}
\end{figure*}



\subsection{Training time speedup}
\label{subsec:comparison_on_training_time_speedup}
\noindent \textbf{Goal}: To minimize the training time while maximizing the flops reduction (FR) and top-1 accuracy. No model retraining is allowed.

\noindent \textbf{Result}: In table \ref{tab:training_time}, we present the Top 1 accuracy, flops reduction and training times for Resnet-50 models on Imagenet. Observe that \methodss saves almost 6 hours with respect to its nearest competitor (GM) with the minimum FR. Further, the model obtained has higher top-1 accuracy. Notably, it saves almost $8$ hours with respect to the original Resnet-50. 

\noindent \textbf{Key Takeaway}: \methodss leads to fast model training.

\begin{table}[h!]
\centering
\begin{adjustbox}{width=\columnwidth,center}
\begin{tabular}{lcccc}
\hline
\textbf{Method} & \multicolumn{1}{c}{\textbf{\begin{tabular}[c]{@{}c@{}}Top-1 \\ Acc (\%)\end{tabular}}} & \textbf{FR} ($\times$) &   \multicolumn{1}{c}{\textbf{\begin{tabular}[c]{@{}c@{}}Training Time\\ (GPU Hours)\end{tabular}}} \\ \hline

Resnet-50         &75.87    & $1.00\times$  & 66.0 \\ 
SFP \cite{he2018soft}   & \textbf{74.61} &  $1.73\times$   &   61.8    \\
GM \cite{he2019filter}    &   74.13  &    $2.15\times$   & 62.2 \\ \rowcolor{Gray}
\methodss \cmmnt{(k=0.03,b=0.3)} &  74.34  & \textbf{2.21}$\times$  & \textbf{51.6}  \\  \bottomrule
\end{tabular}
\end{adjustbox}
\caption{Comparison of training time for Resnet-50 on Imagenet. 1 GPU hour implies $1$ hour of training on 3 Titan Xp GPUs @ 12 GB RAM each. FR indicates flops reduction (higher is better). \methodss saves more than 10 hours with respect to its nearest competitor (GM) while maximizing FR and Top-1 accuracy. Higher Top-1 of SFP is because it is much less compact than other models ($1.73\times$ FR versus $2.15\times$ and $2.21\times$). }
\label{tab:training_time}
\end{table}

\subsection{Comparison on inference time speedup}
\label{subsec:comparison_on_inference_time}
\noindent \textbf{Goal}: To study the effect of \method on inference time.

\noindent \textbf{Result}: In figure \ref{Fig:wallclock_reduction_vs_T}, we plot the inference time speedup versus $t$ for a VGG-16 model trained on CIFAR-10. Indeed, we see that with higher $t$, the speedup increases. Specifically for $t=1$, the model is $4.6\times$ faster.

\noindent \textbf{Key Takeaway}: \method and \methodss speed up inferencing.




\section{Ablation studies}
\label{sec:ablation_studies}
\subsection{On \method for $t$}
\label{subsec:ablation_cup_t}
\noindent \textbf{Goal}: To study the effect of varying $\SC{t}$ in \method.

\noindent \textbf{Result}: In figure \ref{Fig:T_analysis}, We apply \method to $9$ VGG-16 models on CIFAR-10 corresponding to $\SC{t} \sim Uniform(0.7,1.1)$. The final accuracy, parameters reduction, flops reduction and inference time speedup are plotted in fig \ref{Fig:T_analysis}a-d. Like previous studies, we see from fig \ref{Fig:acccuracy_vs_T} that mild compression first leads to an increase (94.17\% from 93.61\%, for $\SC{t}=0.75$) in validation accuracy. However, further compression ($\SC{t}=[0.8-1.1]$ leads to a drop in accuracy.

\noindent \textbf{Key Takeaway}: Higher $t$ leads to more compression.

\subsection{On \methodss for $k,b$}
\label{subsec:ablation_cupss_k_b}
\noindent \textbf{Goal}: To study the effect of varying $\SC{k},\SC{b}$ in \methodss.

\begin{figure}[h!]
\centering
\subfloat[]{\includegraphics[width = 0.23\textwidth]{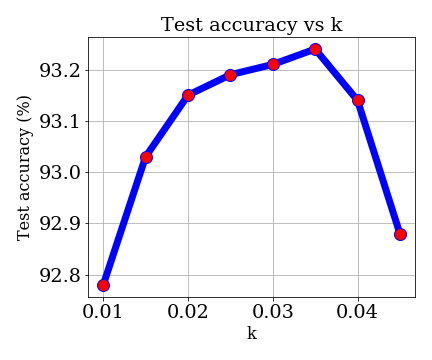} \label{fig:acc_k_vgg}} 
\subfloat[]{\includegraphics[width = 0.23\textwidth]{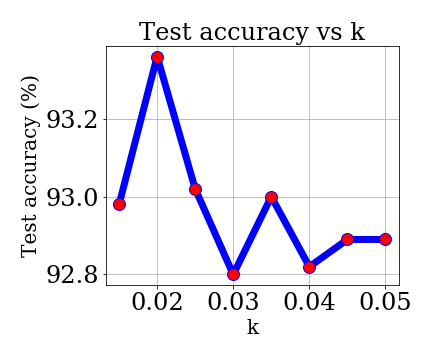} \label{fig:acc_k_resnet}} 
\caption{Validation accuracy on CIFAR-10 versus $\SC{k}$ for compressing (a) VGG-16 and (b) Resnet-56 using \methodss. As intuition suggests, A very fast pruning schedule (high $\SC{k}$) can damage the neural network whereas a very slow pruning rate may lead to incomplete pruning.} 
\label{Fig:acc_vs_k}
\end{figure}

\noindent \textbf{Result}: In fig \ref{Fig:acc_vs_k}, we plot the final accuracy for VGG-16 and Resnet-56 trained on CIFAR-10 versus $k$. Recall that $\SC{k}$ controls the slope of the single-shot pruning schedule introduced in (\ref{eq:single_shot_rate}) while $\SC{b}$ is the offset. We constrain $b=0$. From the figure, we see that for both VGG-16 (fig \ref{fig:acc_k_vgg}) and Resnet-56 (fig \ref{fig:acc_k_resnet}) the ideal value of $\SC{k}$ lies somewhere in the middle of the interval $[0.01,0.05]$. A too small or a too large value of $k$ leads to a larger drop in accuracy. 

\noindent \textbf{Key Takeaway}: A very fast pruning schedule (high $\SC{k}$) can damage the neural network whereas a very slow pruning rate may lead to incomplete pruning.  

\section{Conclusion}
\label{sec:conclusion}
In this paper, we proposed a new channel pruning based method for model compression that prunes entire filters based on similarity. We showed how hierarchical clustering can be used to enable layerwise \textit{non-uniform} pruning whilst introducing only a single hyperparameter. Using various models and datasets, we demonstrated that \method and \methodss achieve the highest flops reduction with the least drop in accuracy. Further, \methodss leads to large savings in training time with only a small drop in performance. We also provided actionable insights on how to set optimal hyperparameters values. In this work, we used a simple yet effective linear trajectory for setting $t$ in \methodss. In future, we would like to extend this to non linear pruning trajectories with goal of further reducing the training time.


\bibliographystyle{aaai_bibst}
\fontsize{9pt}{10pt} \selectfont
\bibliography{aaai}


\appendix
\fontsize{10pt}{11pt} \selectfont

\section{Intuition behind clustering filters}
\label{appendix:intuition_clustering}
In this section we build the intuition behind clustering filters based on features derived from both incoming and outgoing weights. Consider one simple case of a fully connected neural network with $n,m$ filters in layers $l-1$ and $l$, respectively. Here layer $l$ is parametrized by weights $\M{W}^{(l)}\in \mathbb{R}^{m\times n}$ and bias $b^{(l)} \in \mathbb{R}^m$. Within this layer, filter $i$ performs the following nonlinear transformation. 
\begin{equation}
   \label{eq:fcn_op_appendix}
    \V{O}^{(l)}_i = \sigma(\M{W}^{(l)}_{i,:} \V{O}^{(l-1)} + b^{(l)}_i)
\end{equation} Where 
  $\V{O}^{(l-1)}$ is the output from layer $l-1$, $\M{W}^{(l)}_{i,:}, \SC{b}^{(l)}_i$ are the $i^{th}$ row and element of the weight matrix and bias vector respectively. These also constitute the set of incoming weights to filter $i$. Finally $\sigma$ is a non linear function such as RELU. 

\begin{figure}[h!]
\centering
\includegraphics[width = 0.4\textwidth]{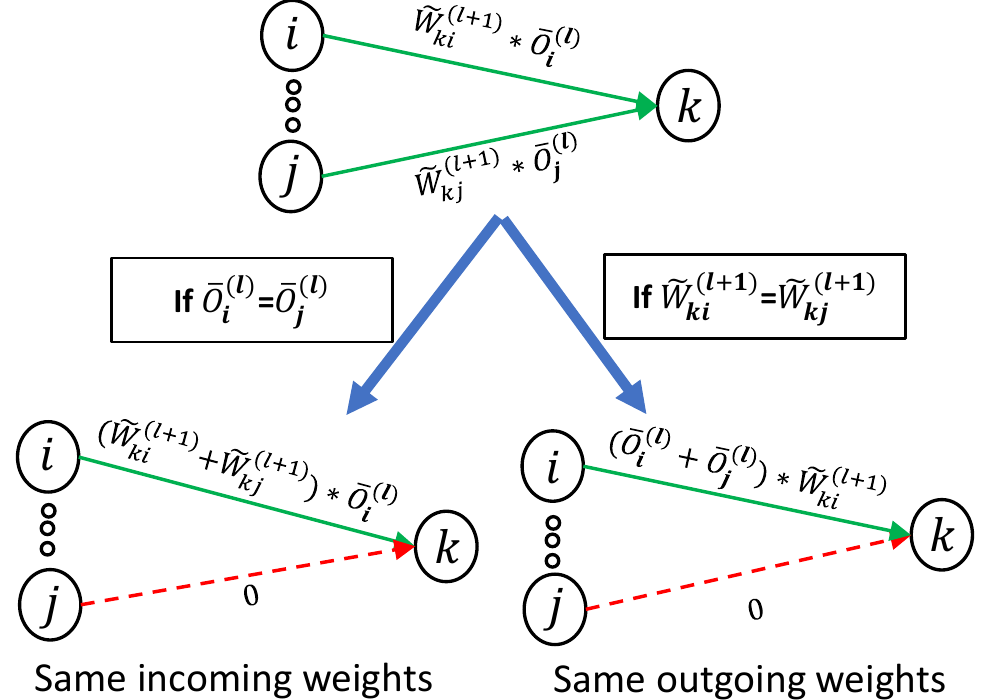} 
\label{fig:contribution}
\caption{Intuition for using input and output weight connections. The top diagram illustrates original neural net structure. The bottom left diagram illustrates the case with same input weights. The bottom right diagram illustrates the case with same output weights}
\label{Fig:intuition}
\end{figure}

Given this notation, a filter $k$ in layer $l+1$ receives the combined contribution from filters $i$ and $j$ equal to $\M{W}^{(l+1)}_{ki} \V{o}^{(l)}_i + \M{W}^{(l+1)}_{kj} \V{o}^{(l)}_j$ where $\M{W}^{(l+1)}_{ki}$ and $\M{W}^{(l+1)}_{kj}$ are the weights from filter $i$ to filter $k$ and from $j$ to $k$, respectively. This is illustrated in the top of figure \ref{Fig:intuition}.

Depending on the input and output weight values, the following two cases can arise 
\begin{itemize}
    
    \item \textbf{If the input weights of filter $i$ and $j$ are same} \cite{srinivas2015data} (i.e., $\M{W}^{(l)}_{i,:} = \M{W}^{(l)}_{j,:}$ and $\SC{b}^{(l)}_i = b^{(l)}_j$ ) or equivalently $\V{o}^{(l)}_i=\V{o}^{(l)}_j$. Then the combined contribution can be provided by a single filter with output connection weight $\M{W}^{(l+1)}_{ki} + \M{W}^{(l+1)}_{kj}$ and the other filter can be pruned. This case corresponds to clustering and pruning similar filters $i$ and $j$ based on input weights and is illustrated in figure \ref{Fig:intuition} bottom left.

    \item \textbf{If the output weights of filters $i$ and $j$ are same} (i.e. $\M{W}^{(l+1)}_{ki}=\M{W}^{(l+1)}_{kj}$) Then combined contribution can be provided by an equivalent filter computing $\V{o}^{(l)}_i+\V{o}^{(l)}_j$ and the other filter can be pruned. This case corresponds to clustering and pruning similar filters $i$ and $j$ based on output weights and is illustrated in figure \ref{Fig:intuition} bottom right.
   
\end{itemize}
Inspired from these idealised scenarios, we propose to identify and prune similar filters based on features derived from both incoming and outgoing weights.

\section{Filter Saliency \& connection to magnitude pruning}
\label{appendix:saliency_and_connection}
\begin{figure}[h!]
\centering
\subfloat[]{\includegraphics[width = 0.25\textwidth]{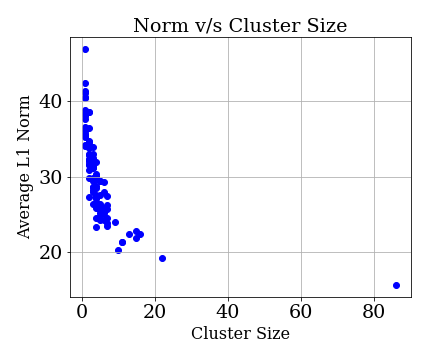} \label{Fig:filter_importance}} 
\subfloat[]{\includegraphics[width = 0.25\textwidth]{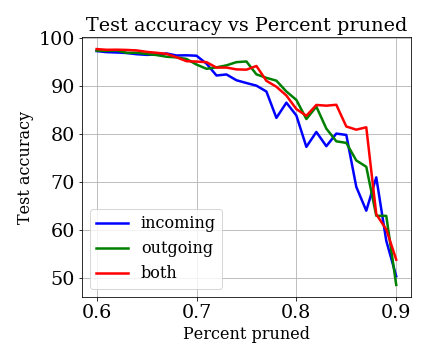} \label{Fig:feature importance}} 
\caption{(a) shows the average $L1$ norm of cluster with size (b) shows the accuracy of the compressed model versus compression.}
\label{Fig:MNIST_analysis}
\end{figure}

\cite{han2015learning} propose to prune individual weight connections based on magnitude thresholding. \cite{li2016pruning} generalize this idea to prune entire filters having a low $L1$ norm of incoming weights. \method generalizes further by using feature similarity as the pruning metric. This metric captures the notion of pruning based on weight magnitude as noted in figure \ref{Fig:filter_importance} where we plot the average $L1$ norm of features for filters in a cluster versus the cluster size. Magnitude pruning operates only at the right end of that figure wherein majority filters have small weights. However, \method operates along the entire axes which means it additionally prunes similar filters that have high weight magnitudes. Hence \method is able to prune a higher number of filters and thus encumbers a lower drop in accuracy for a sufficiently pruned network. This is observed in table \ref{tab:MNIST} where $L1$ and $L2$ based magnitude pruning lead to $13.48 \%$ and $19.04\%$ accuracy  drop (without retraining) versus $8.26 \%$ for \method.

Figure \ref{Fig:feature importance} presents the accuracy for \method using features computed from "incoming", "outgoing" and "both" type  of weight connections. Each point along the x-axis notes the percentage of filters dropped from all layers of model B. This is varied from $[0.6,0.9]$ in steps of 1. The y-axis notes the validation accuracy of the resulting model. It is observed that the combination of input/output features works best.

\section{Comparison with sensitivity driven pruning}
\label{appendix:comparison_sensitivity}
\begin{figure}[t!]
\centering
\includegraphics[width = 0.45\textwidth]{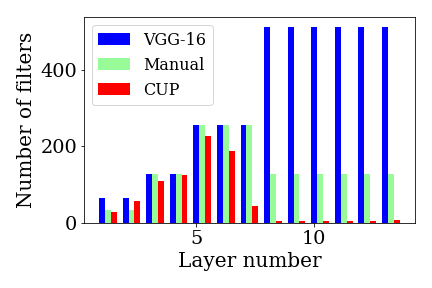} 
\caption{Comparison on number of filters pruned for compressing VGG-16 on CIFAR-10 using \method(T=0.9) versus manual sensitivity analysis in \protect\cite{li2016pruning}. }
\label{Fig:manual_vs_auto}
\end{figure}

In this section, we show similarities in the final model that is automatically discovered by \method versus the manually discovered model in prior works \cite{li2016pruning}. Figure \ref{Fig:manual_vs_auto} compares the number of filters remaining in each layer after compressing VGG-16 on CIFAR-10, through \method (T=0.9) versus that determined through manual sensitivity analysis in \cite{li2016pruning}. Sensitivity analysis involves measuring drop in validation accuracy with pruning filters from layer $l$ while keeping all other layers intact. Layers that are deemed not sensitive are pruned heavily by both methods (layers 8-13) whereas sensitive layers are pruned mildly (layers 1-6).

\section{Line search}
\label{appendix:linesearch}
\subsection{for $t$ in \method} 
The only hyperparameter for compression using the \method framework is the global threshold $\SC{t}$. As a model designer, our goal is to identify the optimal $t$ that yields a model that satisfies the flops budget of the deployment device. In table \ref{tab:linesearch_t_imagenet} we demonstrate the final accuracy achieved by various Resnet variants (trained on Imagenet) for increasing values of $t$. notice how the flops reduction metric (FR) gracefully increases with increasing $t$. This is similar to the observation for compressing VGG-16 on CIFAR-10 as noted in section \ref{sec:ablation_studies} and fig \ref{Fig:flop_reduction_vs_T}.

\begin{table}[h!]
\centering
\begin{adjustbox}{width=0.8\columnwidth,center}
\begin{tabular}{lcc}
\toprule
\multicolumn{1}{l}{\textbf{t}} & \multicolumn{1}{c}{\textbf{FR}} & \multicolumn{1}{c}{\textbf{Accuracy Change (\%)}} \\ 
\multicolumn{1}{c}{\textbf{}}  & \multicolumn{1}{c}{\textbf{($\times$)}}      & \multicolumn{1}{c}{\textbf{Top-1 / Top-5}}   \\ \midrule
\multicolumn{3}{c}{\textbf{Resnet-18 on Imagenet}}                                                                 \\ \midrule
Base                           & 1$\times$                               & 69.88 / 89.26                                \\ \hdashline[3pt/3pt]
0.8                            & 1.75$\times$                               & 68.88 / 88.47                                \\
0.825                          & 1.97 $\times$                               & 68.29 / 88.15                                \\ \midrule
\multicolumn{3}{c}{\textbf{Resnet-34 on Imagenet}}                                                                 \\ \midrule
Base                           & 1$\times$                               & 73.59 / 91.44                                \\ \hdashline[3pt/3pt]
0.60                           & 1.78$\times$                               & 72.73 / 90.91                                \\
0.65                           & 2.08$\times$                               & 71.99 / 90.47                                \\
0.675                          & 2.29$\times$                               & 71.65 / 90.21                                \\
0.70                           & 2.55$\times$                               & 71.15 / 90.08                                \\ \midrule
\multicolumn{3}{c}{\textbf{Resnet 50 on Imagenet}}                                                                 \\ \midrule
Base                           & 1$\times$                               & 75.86 / 92.87                                \\ \hdashline[3pt/3pt]
0.65                           & 2.18$\times$                               & 75.07 / 92.30                                \\
0.675                          & 2.32$\times$                               & 74.73 / 92.14                                \\
0.70                           & 2.47$\times$                               & 74.60 / 92.06                                \\
0.725                          & 2.64$\times$                               & 74.42 / 91.74  \\ \bottomrule
\end{tabular}
\end{adjustbox}
\caption{Line search on $t$ for compressing Resnet variants on Imagenet using \method. Notice that increasing $t$ leads to a graceful increase in flops reduction (FR).}
\label{tab:linesearch_t_imagenet}
\end{table}

\subsection{for $k,b$ in \methodss}
The only hyperparameters for compression using the \methodss framework are $k,b$. From our experience on CIFAR-10, we realize that $k$ needs to be neither too small nor too large. Consequently, we found $k=0.03$ worked well and fixed it for all experiments. For our linesearch presented in table \ref{tab:linesearch_k_b_imagenet}, we vary $b$ over steps of $0.1$. Notice that the reduction in training time for Resnet-50 is much larger than that for other models. This is due to the fact that the original model itself is much larger than its 18 and 34 layer variants. Thus compressing it $2\times$ results in a much larger reduction in flops and training time. As a model designer, our goal is to identify a good $k,b$ that leads to large training time reduction. Having set these values, the designer needs to specify the $tgt\_flops$ (in algorithm \ref{algo1:cupss}) that satisfies the flops budget of the deployment device.

\begin{table}[h!]
\centering
\begin{adjustbox}{width=\columnwidth,center}
\begin{tabular}{lccc}
\toprule
\multicolumn{1}{c}{\textbf{k/b}} & \multicolumn{1}{c}{\textbf{FR}} & \multicolumn{1}{c}{\textbf{Accuracy Change (\%)}} & \textbf{Training} \\
\multicolumn{1}{c}{\textbf{}}    & \multicolumn{1}{c}{\textbf{($\times$)}}      & \multicolumn{1}{c}{\textbf{Top-1 / Top-5}}   &     \textbf{Time (hrs) }         \\ \midrule
\multicolumn{4}{c}{\textbf{Resnet-18 on Imagenet}}                                                                                   \\ \hline
Base                             & 1$\times$                                & 69.88 / 89.26                                & 38.75         \\ \hdashline[3pt/3pt]

0.03/0.3                         & 1.74$\times$                             & 67.38 / 87.86                                & 38.6          \\
0.03/0.4                         & 1.90$\times$                              & 66.86 / 87.37                                & 38.8          \\
0.03/0.5                         & 1.83$\times$                               & 67.24 / 87.59                                & 38.4          \\ \hline
\multicolumn{4}{c}{\textbf{Resnet-34 on Imagenet}}                                                                                   \\ \hline
Base                             & 1$\times$                                & 73.59 / 91.44                                & 44.91         \\ \hdashline[3pt/3pt]

0.03/0.3                         & 1.71$\times$                      & 71.98 / 90.42                                & 39.7          \\
0.03/0.4                         & 2.08$\times$                                & 71.69 / 90.28                                & 39.4          \\ \hline
\multicolumn{4}{c}{\textbf{Resnet 50 on Imagenet}}                                                                                   \\ \hline
Base                             & 1$\times$                                & 75.86 / 92.87                                & 66.00         \\  \hdashline[3pt/3pt]

0.03/0.3                         & 2.20$\times$                                & 74.40 / 91.99                                & 54.48         \\
0.03/0.4                         & 2.16$\times$                                & 74.31 / 92.10                                & 52.45         \\
0.03/0.5                         & 2.21$\times$                                & 74.39 / 91.94                                & 51.58        \\ \bottomrule
\end{tabular}
\end{adjustbox}
\caption{Line search for $k,b$ in compressing Resnet variants using \methodss. We desire a higher FR for an acceptable drop in accuracy. The key observation here is the increasingly larger savings in training time for Resnet 18/34/50.}
\label{tab:linesearch_k_b_imagenet}
\end{table}

\end{document}